\documentclass[11pt]{article}
\usepackage[a4paper,margin=2.5cm]{geometry}
\usepackage[T1]{fontenc}
\usepackage{cite}
\usepackage{amsmath}
\usepackage{newtxtext,newtxmath}
\usepackage{microtype}
\usepackage{graphicx}
\usepackage{textcomp}
\usepackage{xcolor}
\usepackage{booktabs}
\usepackage{float}
\usepackage[font=small,labelfont=bf]{caption}
\usepackage{authblk}
\usepackage{tikz}
\usetikzlibrary{positioning}
\usepackage[hidelinks]{hyperref}

\title{Deep Learning-based 3D Oral Cavity Reconstruction Using 2D Intraoral Images}

\author[1]{Jihun Cho}
\author[2]{Soo-Yeon Jeong}
\author[3]{Eun-Jeong Bae}
\author[1]{Sun-Young Ihm\thanks{Corresponding author.}}
\affil[1]{Dept. of Computer Engineering, Pai Chai University, Daejeon, Korea}
\affil[2]{Division of Software Engineering, Pai Chai University, Daejeon, Korea}
\affil[3]{Dept. of Dental Technology, Bucheon University, Bucheon, Korea}
\affil[ ]{\small\texttt{2161083@pcu.ac.kr}, \texttt{sy.jeong@pcu.ac.kr}, \texttt{baebae@bc.ac.kr}, \texttt{sunnyihm@pcu.ac.kr}}
\date{}

\begin{document}

\maketitle

\begin{abstract}
Oral 3D modelling is one of the most essential stages in dentistry, and many different approaches, such as impression taking and intraoral scanning, are commonly used for this phase, each with notable limitations. Impression taking, which involves placing alginate or silicone material in a tray and inserting it into the patient's oral cavity to form a negative mold, suffers from significant patient discomfort, material deformation errors, and difficulties in storage and transportation. Intraoral scanners, which directly scan oral structures in real time using structured light or laser technology, produce state-of-the-art results but are associated with substantially high equipment costs. To address these limitations, this paper proposes a software-based approach that reconstructs a 3D oral model using only ten 2D intraoral images captured from different angles, requiring no dedicated hardware devices. The proposed method reduces cost, eliminates the need for physical scanning equipment, minimises patient discomfort, and enables automated 3D reconstruction. The model is trained on the publicly available Teeth3DS dataset, comprising 950 upper jaw samples, and employs MobileNetV2 as the image encoder combined with Multi-head Attention for multi-view feature fusion. The proposed model achieves an accuracy of 77.49\%, measured by nearest-neighbor matching with a distance threshold of 0.035. However, predicted vertices tend to concentrate in high-density regions of the ground truth, resulting in uneven point distribution across the reconstructed model.

\end{abstract}

\section{Introduction}\label{sec:intro}

Oral 3D modelling plays a critically important role in dentistry, supporting improved diagnostic accuracy, precision treatment planning, and personalised care across applications such as orthodontics and implant dentistry. It is therefore regarded as an essential stage in modern dental practice. However, conventional oral 3D modelling relies on expensive hardware devices, including intraoral scanners and Cone Beam CT (CBCT), which impose significant cost barriers and limit accessibility in clinical settings. To overcome these constraints, this paper proposes a software-based approach capable of reducing equipment costs, eliminating reliance on physical scanning hardware, minimising patient discomfort, and enabling fully automated 3D reconstruction.

Existing software-based approaches such as Photogrammetry~\cite{westoby2012sfm} and Neural Radiance Fields (NeRF)~\cite{mildenhall2020nerf} require large numbers of images and accurate camera pose estimation, and are designed toward visual scene reconstruction rather than precise geometric coordinate prediction. As such, neither is well-suited for the accuracy demands of dental applications. In contrast, the proposed method requires only ten images taken at predefined angles and directly predicts explicit 3D point cloud coordinates, making it specifically designed for the dental domain.

This paper is an English version of our previous work published at the Korean Multimedia Society Conference~\cite{cho2025korean}.

The remainder of this paper is organised as follows. Section~\ref{sec:related} reviews related work on 3D reconstruction and dental modelling. Section~\ref{sec:method} describes the proposed method, including the dataset, model architecture, and loss function. Section~\ref{sec:experiments} presents experimental results and analysis. Section~\ref{sec:conclusion} concludes the paper with a discussion of limitations and future directions.

\section{Related Work}\label{sec:related}

\subsection{Multi-view 3D Reconstruction}
Multi-view 3D reconstruction has been an active area of research in computer vision. Traditional approaches such as Photogrammetry~\cite{westoby2012sfm} reconstruct 3D geometry by establishing correspondences across multiple images using geometric principles, requiring dense image inputs and accurate camera pose estimation. Neural Radiance Fields (NeRF)~\cite{mildenhall2020nerf} introduced a deep learning-based alternative that represents scenes as continuous volumetric functions mapping 3D coordinates and viewing directions to colour and density, enabling high-quality novel view synthesis from multi-view images. While both methods achieve strong results in general settings, they require large numbers of images and are not designed for direct geometric coordinate prediction, limiting their suitability for domains requiring precise 3D measurement such as dentistry.

\subsection{3D Reconstruction in Dental Applications}
Three-dimensional reconstruction in dentistry has traditionally relied on hardware-based acquisition methods. Intraoral scanners capture oral structures in real time using structured light or laser technology, while Cone Beam CT (CBCT) provides volumetric imaging of tooth and bone structures. Recent studies have explored combining these modalities with deep learning for segmentation and registration tasks~\cite{liu2023dental}. However, these approaches depend on expensive and specialised hardware, limiting their accessibility in clinical settings. In contrast, the proposed method eliminates hardware dependency by reconstructing 3D oral models directly from standard 2D intraoral images.

\subsection{Multi-view Feature Fusion with Attention}
MobileNetV2~\cite{sandler2018mobilenetv2} has been widely adopted as a lightweight and efficient backbone for image feature extraction. The Multi-head Attention mechanism introduced in the Transformer architecture~\cite{vaswani2017attention} has demonstrated strong performance in aggregating information across multiple views, enabling models to capture complementary information from different viewpoints. The combination of CNN-based feature extraction with attention-based multi-view fusion has proven effective in multi-view learning tasks, and forms the basis of the encoder design in the proposed method.

\section{Method}\label{sec:method}

\subsection{Dataset}
The dataset used in this paper is sourced from the publicly available Teeth3DS dataset~\cite{teeth3ds}, consisting of 950 upper jaw and 950 lower jaw mesh files. Each of the 1,900 models contains approximately 150,000 vertices and 300,000 faces on average, with individual teeth classified according to the FDI (F\'{e}d\'{e}ration Dentaire Internationale) numbering system. As this paper focuses on upper jaw reconstruction, only the 950 upper jaw samples are used for training.

\begin{figure}[htbp]
\centering
\begin{tikzpicture}[
    node distance=0.4cm,
    box/.style={rectangle, rounded corners, draw, minimum width=2.3cm,
                minimum height=1.0cm, align=center, font=\small},
    arrow/.style={->, thick}
]
\node[box, fill=gray!15] (raw) {Raw Mesh\\{\tiny $\sim$150K vertices}};
\node[box, fill=teal!20, right=of raw] (fps) {FPS\\{\tiny Centre align + normalise}};
\node[box, fill=gray!15, right=of fps] (sampled) {Sampled\\{\tiny 50K vertices}};
\node[box, fill=teal!20, right=of sampled] (recon) {Poisson\\Reconstruction};
\node[box, fill=purple!20, right=of recon] (render) {Rendering\\{\tiny 10 viewpoints}};
\draw[arrow] (raw) -- (fps);
\draw[arrow] (fps) -- (sampled);
\draw[arrow] (sampled) -- (recon);
\draw[arrow] (recon) -- (render);
\end{tikzpicture}
\caption{Preprocessing pipeline for the Teeth3DS dataset.}
\label{fig:pipeline}
\end{figure}
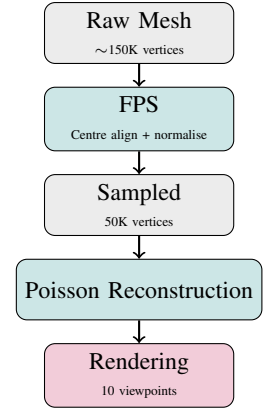

As the raw mesh files are too large to be processed directly, a preprocessing pipeline is applied. Each mesh first undergoes centre alignment and scale normalisation, after which Farthest Point Sampling (FPS)~\cite{qi2017pointnet++} is applied to reduce the number of vertices from approximately 150,000 to 50,000. Unlike random sampling, FPS iteratively selects the vertex furthest from all previously selected vertices, ensuring uniform spatial coverage across the surface. The downsampled vertex set is then used to reconstruct a 3D mesh via Poisson surface reconstruction~\cite{kazhdan2006poisson}. The preprocessing pipeline is illustrated in Fig.~\ref{fig:pipeline}.

\begin{figure}[htbp]
\centering
\includegraphics[width=0.9\linewidth]{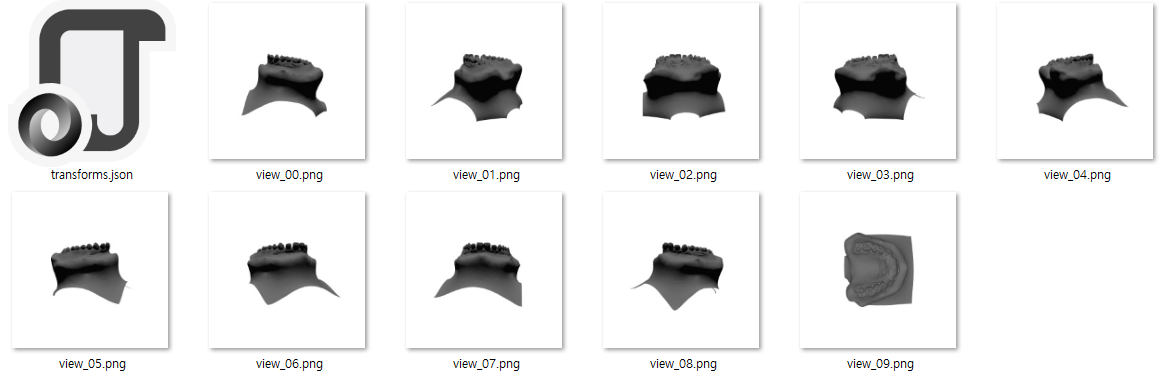}
\caption{Ten fixed viewpoints rendered from a preprocessed upper jaw mesh.}
\label{fig:placeholder}
\end{figure}

Each reconstructed mesh is rendered from ten fixed viewpoints: nine views arranged in a circle at 40° intervals around the mesh, plus one top-down view. The camera is directed toward the mesh centre with a fixed field of view of 45°. This configuration ensures that all surfaces of the oral structure are visible across the ten views, eliminating the need for the model to infer occluded regions. The camera pose for each viewpoint is stored in a transforms.json file and applied consistently throughout training. Example rendered views are shown in Fig.~\ref{fig:placeholder}.

\subsection{Model Architecture}
The proposed model follows an encoder-decoder architecture. The encoder extracts multi-view features from ten input images, which are then passed to the decoder to directly predict 50,000 3D vertex coordinates.

\subsubsection{Encoder}
The encoder takes ten intraoral images captured from different angles as input and produces a single feature vector to be passed to the decoder. Each input sample is represented as a tensor of shape $(B, V, C, H, W)$, where the dimensions are defined as follows:

\begin{table}[H]
\centering
\caption{Input Tensor Dimensions}
\label{tab:input_tensor}
\begin{tabular}{cll}
\toprule
\textbf{Dimension} & \textbf{Description} & \textbf{Value} \\
\midrule
$B$ & Batch size & 2 \\
$V$ & Number of views & 10 \\
$C$ & Colour channels (RGB) & 3 \\
$H$ & Image height & 224 \\
$W$ & Image width & 224 \\
\bottomrule
\end{tabular}
\end{table}

Feature extraction is performed independently for each view using MobileNetV2~\cite{sandler2018mobilenetv2}, with the final classification head replaced by a linear layer that outputs a 512-dimensional feature vector per view, resulting in a tensor of shape $(B, V, F)$ where $F=512$. The encoder is initially frozen for the first 5 epochs to preserve pretrained ImageNet~\cite{deng2009imagenet} features while the decoder stabilises, after which the full network is jointly fine-tuned.

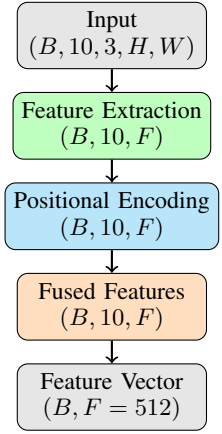
\begin{figure}[htbp]
\centering
\begin{tikzpicture}[
    node distance=0.4cm,
    box/.style={rectangle, rounded corners, draw, minimum width=2.3cm, minimum height=1.0cm, align=center, font=\small},
]
\node[box, fill=gray!20] (input) {Input\\$(B, 10, 3, H, W)$};
\node[box, fill=green!25, right=of input] (feat) {Feature\\Extraction\\$(B, 10, F)$};
\node[box, fill=cyan!25, right=of feat] (pos) {Positional\\Encoding\\$(B, 10, F)$};
\node[box, fill=orange!25, right=of pos] (fused) {Fused Features\\$(B, 10, F)$};
\node[box, fill=gray!20, right=of fused] (out) {Feature Vector\\$(B, F=512)$};
\draw[->, thick] (input) -- (feat);
\draw[->, thick] (feat) -- (pos);
\draw[->, thick] (pos) -- (fused);
\draw[->, thick] (fused) -- (out);
\end{tikzpicture}
\caption{Encoder architecture.}
\label{fig:encoder}
\end{figure}

Positional embeddings are added to each view's feature vector to encode view-specific information, after which multi-head attention~\cite{vaswani2017attention} with 8 heads is applied for view fusion across the ten viewpoints. The attended features are then aggregated via average pooling and refined through an MLP, producing a final output tensor of shape $(B, F)$ where $F=512$, which is passed to the decoder. The encoder architecture is illustrated in Fig.~\ref{fig:encoder}.

\subsubsection{Decoder}
The decoder receives the single feature vector of shape $(B, 512)$ from the encoder and predicts 50,000 3D vertex coordinates as output.

To address the issue of the model producing identical outputs regardless of the input, a Feature Modulator is introduced. The modulator applies a learnable transformation to the feature vector and adds the result at a fixed scale of 0.2, producing a modulated feature that varies with each input and encourages output diversity.

To predict 50,000 vertices, a total of 150,000 dimensions are required, as each vertex consists of three coordinates $(x, y, z)$. An MLP expands the feature dimensions as follows: $512 \rightarrow 1024 \rightarrow 1024 \rightarrow 150{,}000$, where the intermediate layers use ReLU activations and Dropout~\cite{srivastava2014dropout} for regularisation. The output layer directly generates 150,000 values representing the $(x, y, z)$ coordinates of all 50,000 vertices.

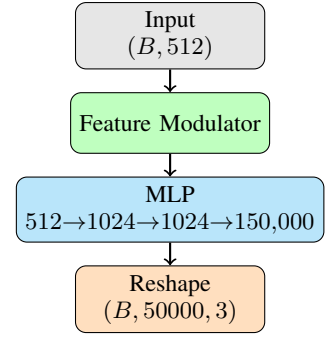
\begin{figure}[htbp]
\centering
\begin{tikzpicture}[
    node distance=0.4cm,
    box/.style={rectangle, rounded corners, draw, minimum width=2.3cm, minimum height=1.0cm, align=center, font=\small},
]
\node[box, fill=gray!20] (input) {Input\\$(B, 512)$};
\node[box, fill=green!25, right=of input] (mod) {Feature\\Modulator};
\node[box, fill=cyan!25, right=of mod] (mlp) {MLP\\$512{\to}1024{\to}1024{\to}150{,}000$};
\node[box, fill=orange!25, right=of mlp] (reshape) {Reshape\\$(B, 50000, 3)$};
\draw[->, thick] (input) -- (mod);
\draw[->, thick] (mod) -- (mlp);
\draw[->, thick] (mlp) -- (reshape);
\end{tikzpicture}
\caption{Decoder architecture.}
\label{fig:decoder}
\end{figure}

The output is reshaped from a 150,000-dimensional vector into a tensor of shape $(B, 50000, 3)$. A tanh activation is applied to constrain all coordinate values to the range $[-1, 1]$. Finally, a Scale Predictor adjusts the overall scale of the predicted coordinates, while a Coord Scale Predictor independently adjusts the scale along each of the $x$, $y$, and $z$ axes. The decoder architecture is illustrated in Fig.~\ref{fig:decoder}.

\subsection{Loss Function}

Two loss functions are combined in this paper, as defined in Equation~\ref{eq:loss}:

\begin{equation}
\mathcal{L} = \alpha \cdot \mathcal{L}_{Chamfer} + \beta \cdot \mathcal{L}_{L1}
\label{eq:loss}
\end{equation}

L1 Loss, which directly compares the predicted and ground truth coordinates by computing their mean absolute difference, is assigned a higher weight in the early stages of training to focus on fine-grained positional alignment. Chamfer Distance, which computes the nearest-neighbour distances bidirectionally between the predicted and ground truth point sets, is assigned a higher weight in the later stages to capture overall structural accuracy. The epoch-wise weight scheduling is summarised in Table~\ref{tab:weight_schedule}.

\begin{table}[htbp]
\centering
\caption{Loss Weight Scheduling by Epoch}
\label{tab:weight_schedule}
\begin{tabular}{ccc}
\toprule
\textbf{Epoch} & \textbf{$\alpha$ (Chamfer)} & \textbf{$\beta$ (L1)} \\
\midrule
0 -- 4  & 0.3 & 0.7 \\
5 -- 14 & 0.6 & 0.4 \\
15+     & 0.8 & 0.2 \\
\bottomrule
\end{tabular}
\end{table}

\section{Experiments}\label{sec:experiments}

\subsection{Experimental Setup}
All experiments were conducted on a single NVIDIA RTX 5070 GPU. The detailed training configuration is summarised in Table~\ref{tab:setup}.

\begin{table}[htbp]
\centering
\caption{Experimental Setup}
\label{tab:setup}
\begin{tabular}{ll}
\toprule
\textbf{Configuration} & \textbf{Value} \\
\midrule
GPU & NVIDIA RTX 5070 (12GB VRAM) \\
Python & 3.12.9 \\
PyTorch & 2.8.0 \\
CUDA & 12.8 \\
Optimizer & AdamW~\cite{loshchilov2019adamw} \\
Initial learning rate & 0.001 \\
Weight decay & $1 \times 10^{-4}$ \\
LR scheduler & CosineAnnealingLR ($T_{max}=30$) \\
Batch size & 2 \\
Epochs & 50 \\
Early stopping patience & 15 \\
Vertex count & 50,000 \\
\bottomrule
\end{tabular}
\end{table}

\subsection{Results}
\begin{figure}[htbp]
\centering
\includegraphics[width=0.8\linewidth]{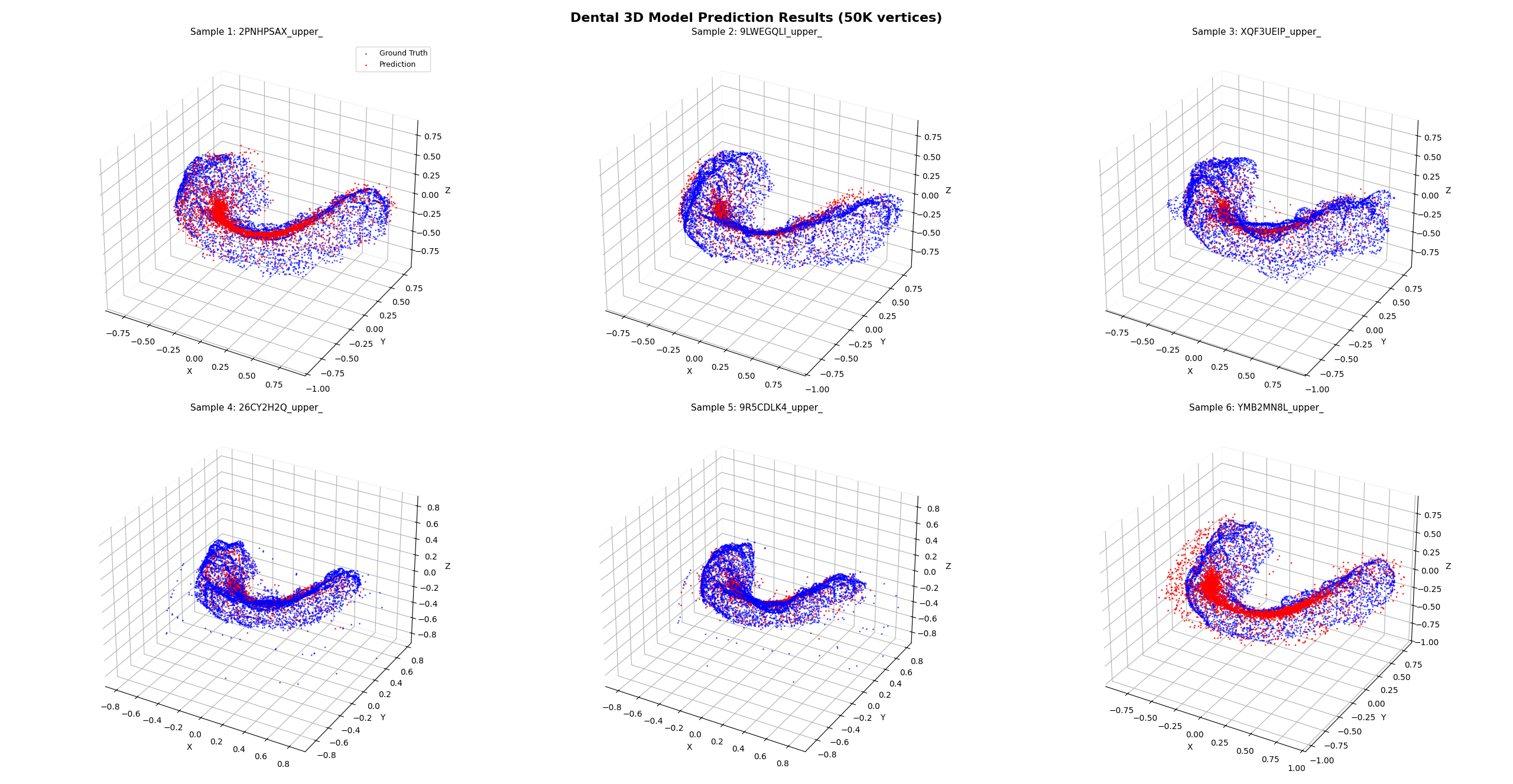}
\caption{Prediction results on six upper jaw samples. Blue: Ground Truth, Red: Prediction.}
\label{fig:results}
\end{figure}

The proposed model achieved an accuracy of 77.49\%, converging at epoch 47 of 50. The final Total Loss was 0.133597, with a Chamfer Loss of 0.090422 and an L1 Loss of 0.306298. As shown in Fig.~\ref{fig:results}, predicted vertices are clustered in high-density areas of the ground truth, resulting in uneven point distribution across the reconstructed model.

\subsection{Analysis}
The clustering of predicted vertices in high-density regions can be attributed to the nature of the Chamfer Distance loss function. Chamfer Distance computes the nearest-neighbour distance from each predicted point to the ground truth, and vice versa. As a result, the model is rewarded for placing predicted vertices near any ground truth point, regardless of whether that region is already densely covered. This encourages the model to concentrate predictions in areas where ground truth vertices are densely distributed, as doing so minimises the overall loss more efficiently. Consequently, sparse regions of the oral structure receive fewer predicted vertices, leading to the uneven distribution observed in the results.

\section{Conclusion}\label{sec:conclusion}
This paper proposes a supervised deep learning framework for 3D oral cavity reconstruction from ten fixed-angle 2D intraoral images, eliminating the need for expensive hardware devices such as intraoral scanners or CBCT equipment. The proposed model, based on MobileNetV2 and Multi-head Attention, directly predicts 50,000 3D vertex coordinates from multi-view images, achieving an accuracy of 77.49\%, measured by nearest-neighbor matching with a distance threshold of 0.035.

However, the results reveal a limitation in the use of Chamfer Distance as the primary loss function. Predicted vertices tend to concentrate in high-density regions of the ground truth, resulting in uneven point distribution across the reconstructed model. This uneven distribution reduces the practical utility of the reconstruction in clinical dental applications.

Future work will address this limitation by introducing an improved loss function designed to enforce more uniform vertex distribution, aiming to produce geometrically accurate and clinically applicable 3D oral models.

\section*{Acknowledgment}
This work was supported by the Institute of Information \& Communications Technology Planning \& Evaluation (IITP) grant funded by the Korea government (MSIT) (IITP-2025-RS-2022-00156334, 50\%), and by the National Research Foundation of Korea (NRF) grant funded by the Korea government (MSIT) (No. 2021R1C1C2011105, 50\%).

\end{document}